\documentclass[letterpaper, 10 pt, conference]{ieeeconf}

\IEEEoverridecommandlockouts
\usepackage{cite}
\usepackage{amsmath,amssymb,amsfonts}
\usepackage{algorithmic}
\usepackage{graphicx}
\usepackage{textcomp}
\usepackage{xcolor}
\usepackage{balance}
\usepackage{comment}
\usepackage[ruled,vlined]{algorithm2e}
\usepackage{caption}
\usepackage{array}
\usepackage{subcaption}
\usepackage{arydshln}
\usepackage{tensor}
\newcommand{\eg}{\emph{e.g.},}
\newcommand{\ie}{\emph{i.e.},}

\usepackage{hyperref}
\usepackage{float}
\usepackage{rotating}
\usepackage{nicematrix}  
\usepackage{mathtools}
\usepackage{pifont}
\usepackage{cuted}
\usepackage{dblfloatfix}

\definecolor{LightCyan}{rgb}{0.88,1,0.88}
\definecolor{linear_color}{RGB}{220,223,240}
\definecolor{gray_bbox_color}{RGB}{243,243,244}

\definecolor{rebuttal}{rgb}{0,0,1}

\def\eqref#1{Eq.~(\ref{#1})}

\DeclareMathOperator*{\argmin}{argmin}

\usepackage{fancybox}
\usepackage{hhline}
\usepackage{multirow}
\usepackage{color,soul}

\newcommand{\ForceSym}{     \vec{u}             }

\newcommand{\ForceMod}{ F }
\newcommand{\ODmod}{ \mathcal{OD} }

\newcommand{\OurMethod}{Shape-Space Deformer}

\newcommand{\NovelMethodNumEpochs}{$500$}

\newcommand{\ExperimentThreeNE}{
    Random
}

\newcommand{\ExperimentThree}{\emph{\ExperimentThreeNE}}

\newcommand{\oneWayChamferSym}[2]{ \mathcal{CD}({#1}{\rightarrow}{#2}) }
\newcommand{\oneWayChamferDef}[2]{
    \text{mean}
    \big(
        \min_{\mathbf{\MakeLowercase{#2}} \in #2} {
            \| \mathbf{\MakeLowercase{#1}} - \mathbf{\MakeLowercase{#2}} \|
        }^{2}
    \big)
}

\newcommand{\twoWayChamferSym}[2]{ \mathcal{CD}({#1}{\leftrightarrow}{#2}) }
\newcommand{\twoWayChamferDef}[2]{
    \big[ \mathcal{CD}({#1}{\rightarrow}{#2}) + \mathcal{CD}({#2}{\rightarrow}{#1}) \big] \div 2
}

\usepackage{booktabs, makecell}

\begin{document}


\title{\LARGE \bf Shape-Space Deformer: Unified Visuo-Tactile Representations for Robotic Manipulation of Deformable Objects}

\author{
    Sean M. V. Collins$^{1,2}$, Brendan Tidd$^{1}$, Mahsa Baktashmotlagh$^{2}$, Peyman Moghadam$^{1,2}$
    \thanks{
        $^{1}$ CSIRO Robotics, Data61, CSIRO, 
        Australia. 
        E-mails: {\tt\footnotesize \emph{firstname.lastname}@csiro.au}
    }
    \thanks{
        $^{2}$ School of Electrical Engineering and Computer Science (EECS), The University of Queensland, St Lucia, Australia.
    }
    \thanks{ $^{*}$ This work was supported by the CSIRO's Data61 Embodied AI Cluster.}
}
 

\bstctlcite{IEEEexample:BSTcontrol}

\maketitle

\begin{abstract}
Accurate modelling of object deformations is crucial for a wide range of robotic manipulation tasks, where interacting with soft or deformable objects is essential.
Current methods struggle to generalise to unseen forces or adapt to new objects, limiting their utility in real-world applications.
We propose \OurMethod, a unified representation for encoding a diverse range of object deformations using template augmentation to achieve robust, fine-grained reconstructions that are resilient to outliers and unwanted artefacts.
Our method improves generalization to unseen forces and can rapidly adapt to novel objects, significantly outperforming existing approaches.
We perform extensive experiments to test a range of force generalisation settings and evaluate our method's ability to reconstruct unseen deformations. Our results demonstrate significant improvements in reconstruction accuracy and robustness.
Our approach is suitable for real-time performance, making it ready for downstream manipulation applications. 
\\
\\
\end{abstract}

%
%

\section{Introduction} \label{sec:intro}
Manipulation of deformable objects is a fundamental but challenging problem for deploying robots in human-centered environments ranging from industrial and services to household and surgical robotics. However, existing robot manipulation research mostly focuses on rigid object types~\cite{celemin2022interactive, weng2023neural}. The ability to accurately predict and reconstruct object deformations subject to external forces plays a critical role in planning safe interactions in the real world. Unlike rigid object manipulation, deformable objects are highly dynamic and require high dimensional state representation.
    
    Recently, Neural Fields (NF), have emerged as a powerful alternative method for modelling and representing scenes for robotic applications due to their compactness, continuity, and scalability.
    These neural representations (NeRF, SDF, and other forms), have shown great potential in robotics applications such as SLAM~\cite{Ortiz_2023_iSDF2022, Mao_2024}, navigation~\cite{adamkiewicz2022vision, Kwon_2023_CVPR},  object registration~\cite{regnf2024}, and manipulation~\cite{rashid2023language, weng2023neural}. Notably, several methods now employ neural fields for representing deformable for diverse tasks including human body parts modelling \cite{Park_2021_ICCV}, dynamics scences~\cite{Pumarola_2021_CVPR}, and garment representation~\cite{Chi_2021_ICCV, Aggarwal_2022_ACCV}. 
    
    Most recently, VIRDO and its variant~\cite{Wi_2022,wi2023virdo++} extended these compact latent representations to effectively integrate tactile and visual information for robotic manipulation.
    Specifically, VIRDO uses a two-stage network to build a joint representation for visuo-tactile state estimation subject to external forces. They achieve this by first modelling shape representation for nominal (undeformed) shapes, before training on deformations.
    Despite their impressive results, VIRDO suffers from poor generalization to unseen forces (\ie{} few or zero-shot settings),
    thus limiting their broader applicability since re-training is inevitably needed for unseen forces applied to the same object categories. This limitation is highlighted through our experiments utilizing the VIRDO dataset, where a subset of force samples are deliberately omitted from training to test model robustness.   
    To address this challenge, we introduce the Shape-Space Deformer, a new visuo-tactile representation method that learns a unified shape representation for both deformed and nominal shapes subject to external forces. 
    \begin{figure}[t!]
        \centering
        \includegraphics[width=0.5\textwidth]{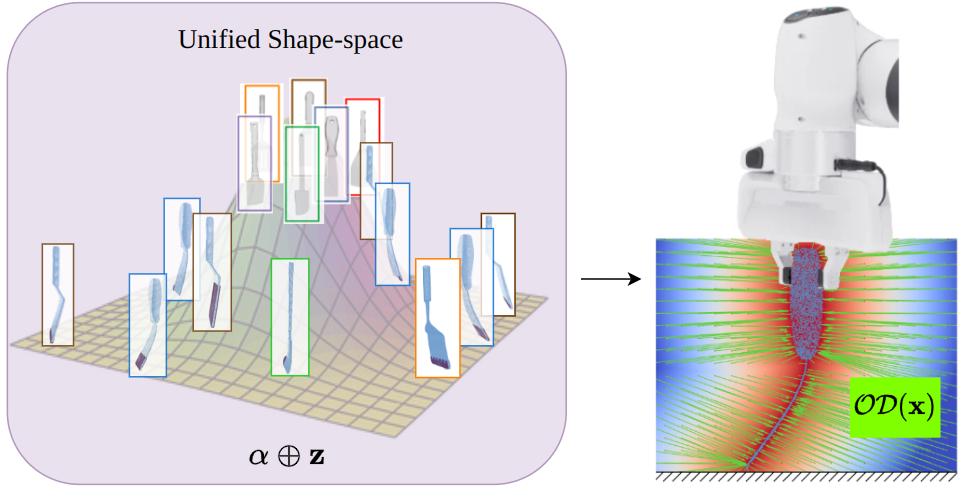}
        \caption{We present a unified shape representation for learning how objects $\alpha$ deform given a set of encoded forces $\mathbf{z}$. Our method generalises to unseen forces and new objects when supplied with only a few example deformations.}
        \label{fig:figure_one}
    \end{figure}
To summarize, our main contributions are: 
\begin{itemize}
\item A novel \textit{unified latent representation} that effectively encodes a diverse range of object deformations to enhance model learning and generalisation.  
\item An explicit neural field rendering technique that augments a template shape, achieving fine-grained reconstructions, ensuring a robustness to outliers and stray artifacts, and exhibiting real-time performance.
\item Extensive experiments to demonstrate generalisation to unseen forces and adaptation to new object types with minimal training data. We display significant improvements over baselines in deformation reconstruction and applicability for the real-world.
\end{itemize}
\begin{figure*}[th!]
    \centering
    \includegraphics[width=0.80\textwidth]{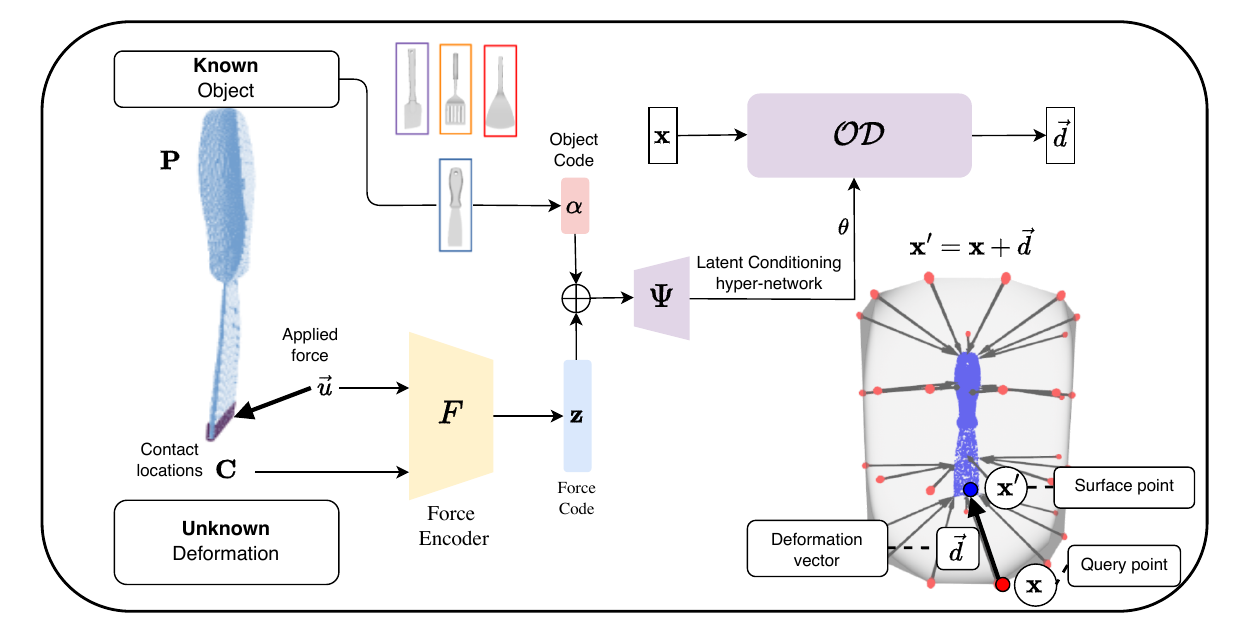}
    \caption{An overview of our Shape-Space Deformer Network. Given a known object class, contact locations, and applied force, our method determines the corresponding deformation field. We create a unified representation of several objects and their deformation states and generate a surface reconstruction by explicitly learning the neural field from a template shape. }
    \label{fig:arch}
\end{figure*}

\section{Related Work} \label{sec:rel_work}

\subsection{Neural Fields for Deformable Objects}
    Beginning on the seminal work of Nerfie~\cite{Park_2021_ICCV}, numerous studies have employed neural fields for dynamic scene reconstruction, particularly on modelling sequences of time-varying, non-rigid scenes \cite{Pumarola_2021_CVPR, Park_2021, Peng_2021_ICCV, Yan_2023_CVPR, yan2023od, Park_2023_CVPR}. 
    These methods typically learn a canonical template through neural fields, while a separate neural deformation field deforms coordinates between each scene frame and the template, establishing correspondences and enabling the representation of the scene at different time steps.
    DeformNet~\cite{Li_2024} further refines these latent representations by employing separate embeddings for both deformation and appearance. This distinction is grounded in the observation that certain deformations are better captured by appearance changes (\eg{} in objects with diverse colouring), whereas others are more effectively detected through visible motion (\eg{} uniformly coloured objects). 
    Deformable neural fields have found applications across various domains, including the modelling of moving human bodies~\cite{Park_2021_ICCV, Peng_2021_ICCV, Weng_2022_CVPR}  and in medical anatomy~\cite{wang2022neural,shi2023colonnerf}. 
    %
    %
    This task has also been extended to scenes with multiple interacting objects, where the scene is decomposed by representing each object with a separate neural field. Various techniques are then employed to combine these individual components into a cohesive scene~\cite{driess2023learning, Song_2023_ICCV}.
    %
    DIF-Net~\cite{Deng_2021_CVPR} learns a template signed distance function (SDF) and a volumetric deformation function using MLPs, which provides dense correspondences between generated SDFs for shapes within a diverse range of categories.
    %
\subsection{Multimodal Visuo-Tactile Representations}
    %
    %
    %
    %
    %
    Visuo-tactile inputs comprise diverse modalities, with useful feature information for modelling contact forces and deformations.
    Several recent works have focused on integrating tactile data with other sensory modalities, particularly visual, to enhance the understanding of objects for manipulation, and often presenting generative applications.
    %
    Touching a NeRF~\cite{zhong2023touching} uses a NeRF to render RGB-D images of objects, which are then encoded and input to a generator network to infer an output tactile image. The RGB-D image and either the inferred tactile image or ground truth are then fed to a discriminator network, to distinguish real and fake images.
    TaRF~\cite{Dou_2024_CVPR} explores fusing visuo-tactile input with whole scenes, presenting a neural field for tactile signals, alongside a NeRF for scene representations.
    %
    %
    VIRDO~\cite{Wi_2022} targets the challenging problem of modelling the deformation of objects based on observations of deformations resulting from applied contact forces without knowing object physical/material parameters. VIRDO++\cite{wi2023virdo++} then extends and applies to state-estimation and dynamics prediction using an action module to future boundary conditions. 
\section{Methodology}
    In this section, we begin by elaborating on the architecture of our main network, which is designed to represent the geometric variations across all observed $n \times m$ deformations.
    Each object is parameterized by embeddings $\alpha$, and the corresponding deformations are encoded by force codes $\mathbf{z}$.
    Fig.~\ref{fig:arch} depicts the proposed pipeline: starting from a known object class, contact locations and applied force, a shared shape representation is learned that encompasses all objects and their deformations, which are reconstructed by transforming a template shape.
    The initial sections explain the roles of these embeddings and force codes, setting the stage for explaining how these condition the behaviour of $\mathcal{OD}$ to model each specific shape instance.
    This is followed by a detailed description of our $\mathcal{OD}$ module and how it is able to capture geometric properties and deformation behaviours of the objects.
    %
\begin{figure}[htbp]
\centering
\includegraphics[width=1\linewidth]{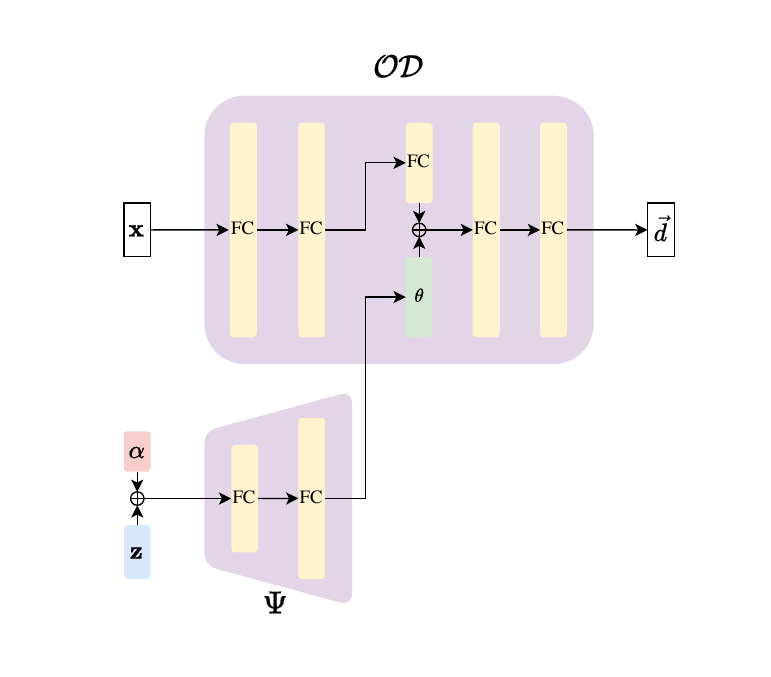}
\caption{The architecture of the hyper-network. Our unified representation takes an object code and latent force vector to condition a single policy trained on all shape types and deformation examples to predict the augmentation to be applied to a cylindrical shape template.}
\label{fig:hypernet}
\end{figure}
\subsection{Unified Latent Representation} \label{subsec:latent}
    Our main network $\mathcal{OD}$ represents the geometry of all object deformations from a given dataset.
    To condition $\mathcal{OD}$ according to a particular object and deformation, we parameterise the $n$ objects with embeddings $\alpha$, and encode the $m$ deformations for each object with force codes $\mathbf{z}$.
    The conditioning of $\mathcal{OD}$ to represent multiple objects and deformations is referred to as a hyper-network, which is an MLP decoder, taking a latent conditioning vector as input and outputting a set of parameters for a main network \cite{NEURIPS2020_53c04118, Deng_2021_CVPR}.
    %

    \noindent \textbf{Hyper-Network $\Psi$}. We employ a hyper-network $\Psi$ that takes object codes $\alpha$ and a force code $\mathbf{z}$ and generates the parameters of the MLP for our \OurMethod{} network. Each object category ($\alpha_{i}$) is represented with a latent object code $\alpha_{i} \in \mathbb{R}^{l}$. We initialise the object codes from a normal distribution. 
    To generate the force code, $\mathbf{z}_{ij} \in \mathbb{R}^{d}$, we follow the force encoder architecture similar to VIRDO~\cite{Wi_2022}.  For each deformation instance $j$ of an object, visuo-tactile input data is provided in the form of a reaction force $\vec{u}_{ij}$ and a set of contact locations $\mathbf{C}_{ij}$. The latent object code and force network are optimised jointly with the parameters of the hyper-network $\Psi$ and main network $\mathcal{OD}$. Fig.~\ref{fig:hypernet} illustrates this architecture. 
    
    Unlike VIRDO~\cite{Wi_2022}, which represents the modelling of a shape as a deformation of a fixed, nominal object (\ie{} template), our method models deformation for a given shape as a continuous function over a learned (not fixed) template.
    We achieve this by first concatenating $\alpha$ and $\mathbf{z}$ into the hyper-input for $\Psi$, which decodes this into an output $\theta$ (Eq.~\ref{eq:hyperOutput}). The output $\theta$ is then joined with output from a hidden layer to condition the main network $\mathcal{OD}$ (Fig.~\ref{fig:hypernet}). We use the shorthand $\mathcal{OD}_{ij}$ to refer to the main network being conditioned on $\theta_{ij}$ from (object~$i$, deform~$j$). This differs from VIRDO\cite{Wi_2022}, which first pre-trained a nominal shape network, fixed the learned object codes, and then used the shape network throughout for deformation modelling. Our network dynamically learns the object code that represents the entire object category (deformed and undeformed).
    \begin{equation}
    \label{eq:hyperOutput}
            \theta_{ij} = \Psi{}(\alpha_{i}, \mathbf{z}_{ij}).
    \end{equation}
\subsection{Deformation Learning}\label{subsec:defLearning}
    %
    To represent geometry, our \OurMethod{} module $\mathcal{OD}$ learns a deformation field, mapping query points in 3D space to an object's surface. 
    For an input query point $\mathbf{x}$, $\mathcal{OD}$ returns a displacement vector $\vec{d}$, between $\mathbf{x}$ and the closest point $\mathbf{p}$ on the object surface $\mathbf{P}$ (Eq.~\ref{eq:OD_defn}). 
    Consequently, the surface point is simply this vector added to $\mathbf{x}$ (Eq.~\ref{eq:OD_finalOD}).
    \begin{equation}
    \label{eq:OD_defn}
        \mathcal{OD}(\mathbf{x}) = \argmin_{ \mathbf{p} \in \mathbf{P} }{\big\{ \| \mathbf{p} - \mathbf{x} \| \big\}}; \quad \mathcal{OD} : \mathbb{R}^{3} \rightarrow \mathbb{R}^{3}, 
    \end{equation}    
    \begin{equation}
    \label{eq:OD_finalOD}
    {\mathbf{x}^ \prime} = \mathbf{x} + \mathcal{OD}(\mathbf{x}).
    \end{equation}  
    %
    %
    %
\subsection{Getting to the Surface}
\begin{figure}[t!]
            \centering\includegraphics[width=0.48\textwidth]{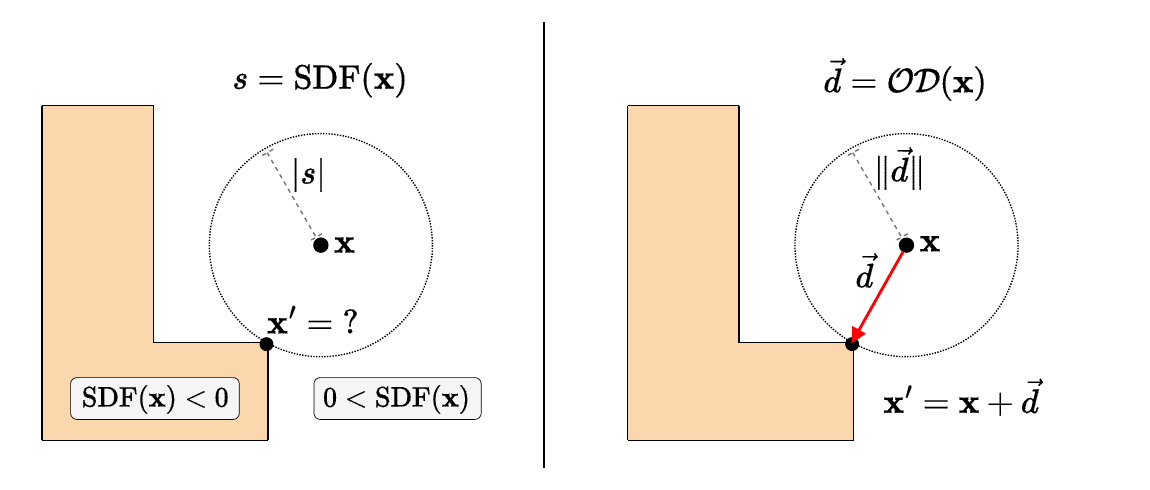}
            \caption{
            Left: SDFs only describe distance from a surface.
            Right: Our model $\mathcal{OD}$ focuses on ``getting to the surface''.
            }
            \label{fig:heuristic1}
        \end{figure}

    In this section, we explain the application of $\mathcal{OD}$'s vector field both for representing surface geometry and deformation from applied forces.
    %
    %
    The input for $\mathcal{OD}$ is a Cartesian point $\mathbf{x} \in \mathbb{R}^{3}$, and the output is a vector $\vec{d} \in \mathbb{R}^{3}$ to apply on this point to express exactly where the closest surface point lies.
    SDF based method such as \cite{Deng_2021_CVPR, Wi_2022} output a scalar $s \in \mathbb{R}$, only indicating how far away the nearest surface point is, that is, the sphere in which the point lies, but not the direction (Fig.~\ref{fig:heuristic1}).
    A salient observation is that the `density' of points on an object surface is much less than the off-surface volume. 
    Optimising $\mathcal{OD}$ is a \textbf{regression} problem, taking coordinates from the \emph{voluminous} domain of $\mathbb{R}^{3}$ and mapping them specifically onto an object \emph{surface}.
    The intent is that all outputs of $\mathcal{OD}$ relate explicitly to the surface, in contrast with other neural fields (such as SDFs), which do not explicitly learn the surface.
    SDFs could be deemed as a \textbf{classifier}, assigning points from the same domain into three groups: the \emph{volume} outside of an object, the object surface, or inside the object \emph{volume}.
    Surface details are important to deformation modelling because this is also where the contact locations are. 
    By concentrating on the surface, our method is better able to synthesise these details with the contact locations encoded by $F$.
    %
        %
        %
    \subsection{Explicit Shape Rendering} \label{subsec:rendering}
        %
        The objective of mapping all of $\mathbb{R}^{3}$ to exclusively surface points is advantageous because it removes the need to perform extraction algorithms on high-resolution grid samples to find (approximate) surface points.
        Instead, Equation~\ref{eq:OD_finalOD} provides an \emph{explicit} formula to yield surface points.
        To render a mesh, we use $\mathcal{OD}_{ij}$ to displace the vertices of a template cylinder mesh into the particular (shape, deformation) instance.
        A cylinder mesh is initialised to already have connected triangle faces.
        We take all of its vertices as input to warp, returning a mesh deformed into the desired shape. 
\subsection{Objective Function} \label{subsec:losses}
    %
    %
    %
    Finally, our total loss ($\mathcal{L}_\mathcal{OD}$) can be presented as: 
    
    \begin{equation}
    \mathcal{L}_\mathcal{OD} = \lambda_{shape} \mathcal{L}_{shape} + \lambda_{reg} \mathcal{L}_{reg},
    \end{equation}
    which is composed of two main terms: a shape loss that focuses optimisation on accurate reconstructions of the nominal and deformed shapes ($\mathcal{L}_{shape}$), and a regularisation loss for the neural parameters of our method ($\mathcal{L}_{reg}$). $\mathcal{L}_{shape}$ is represented as:  
    \begin{equation}
            \mathcal{L}_{shape} = \mathcal{L}_{vec} + \mathcal{L}_\mathcal{CD}.
    \end{equation}
    
    $\mathcal{L}_{vec}$ is the mean $L_2$ loss between the estimated closest points to $\mathbf{x}$ (using Eq.~\ref{eq:OD_finalOD}) and the ground truth $\mathbf{p}^{*}$ (Eq.~\ref{eq:vec_loss}). We take the mean over a number of on-and-off-surface sample points within a unit cube $\mathbf{x} \in \Omega \subset[-1, +1]^{3}$.
    %
    %
    \begin{equation}
    \label{eq:vec_loss}
            \mathcal{L}_{vec} = \frac{1}{| \Omega |} \sum_{\mathbf{x} \in \Omega} 
                \big\| (\mathbf{x} + \mathcal{OD}(\mathbf{x})) - \mathbf{p}^{*} \big\|^{2}.
    \end{equation}
    
    $\mathcal{L}_\mathcal{CD}$ is the mean bi-directional $L_2$ Chamfer Distance between a point cloud reconstruction of a shape using $\mathcal{OD}$, and the ground truth point cloud $\mathbf{P}^{*}$.
    %
    \begin{equation}
    \label{eq:CDdef}
            \mathcal{L}_\mathcal{CD} = \mathcal{CD}(\mathbf{P}{\leftrightarrow}\mathbf{P}^{*}),
    \end{equation}    
    where $\twoWayChamferSym{A}{B} := \twoWayChamferDef{A}{B}$ and $\oneWayChamferSym{A}{B} := \oneWayChamferDef{A}{B} \quad \forall\mathbf{a}\in A$.
    We use two regularisation losses, $\mathcal{L}_{\alpha}$ to promote the latent shape space to have low variance and mean close to zero (Eq.~\ref{eq:emb_loss}), and the loss $\mathcal{L}_{\mathbf{w}}$ for regularisation of the weights of $\ODmod$.
    The two losses take the mean-squared sum of the shape codes and $\ODmod$ weights, respectively, encouraging the parameters to have a zero mean and low variance.
    \begin{equation}
    \label{eq:emb_loss}
        \mathcal{L}_{reg} = \mathcal{L}_{\alpha} + \mathcal{L}_{\mathbf{w}},
    \end{equation}
    where $\mathcal{L}_{\alpha} = \text{mean}\big(\| \alpha \|^{2})$ and $\mathcal{L}_{\mathbf{w}} = \text{mean}\big(\|\mathbf{w}\|^{2})$ for $\forall \alpha$ shape codes, and weights $\mathbf{w}$ from $\mathcal{OD}$.
\section{Experiments} \label{sec:experiments}
    \subsection{Experimental Design}
    In evaluating our method, we have three categories of experiments that reinforce our contributions.
    First, \textbf{shape reconstruction with known reconstruction} experiments show we can produce high quality reconstructions of nominal and deformed shapes, similar to \cite{Wi_2022}.
    Secondly, \textbf{force generalisation} experiments investigate the effect of withholding forces during training and show we can generalise to a variety of forces by reconstructing shapes subjected to the withheld forces (\emph{Random}, \emph{Lowest}, \emph{Highest}, and \emph{Direction}, described in Section~\ref{subsec:force_gen}).
    %
    Finally, \textbf{object generalisation} experiments demonstrate that we can extend our network to new shapes with limited deformation examples (\emph{No deformed} and \emph{k deformed}, described in Section~\ref{subsec:obj_gen})).
    We use the dataset introduced in VIRDO~\cite{Wi_2022}, comprised of $n=6$ kitchen utensils (\eg{} spatulas, spoons), each with a nominal shape and $m=24$ deformed shape samples.
    Each deformed sample has the deformed point cloud, contact locations and applied force vector.
\subsection{Training and Implementation Details}
    We choose a size of $l=6$ for the object codes, \ie{} $\alpha\in\mathbb{R}^{6}$, with these parameters randomly initialised from a normal distribution $\alpha\sim\mathcal{N}(0, 0.1^2)$ before training.
    For the force codes, we select a size of $d=12$, \ie{} $z \in \mathbb{R}^{12}$, which is consequently the output size of the force module $\ForceMod$.
    The 1.0M parameters from $\alpha$, $\ForceMod$ and $\ODmod$ are jointly learnt through \NovelMethodNumEpochs{} epochs of training, using the Adam optimisation algorithm \cite{kingma_2014_Adam}, with learning rates of $1\times10^{-4}$ for $\alpha$, and $5\times10^{-5}$ for $\ODmod$ and $\ForceMod$.
    We select the loss weightings 
    $\lambda_{shape} = 5\times10^{6}$ and
    $\lambda_{reg} = 10^{2}$. 

    Coordinates for ground-truth objects are normalised to be within $\pm{1}$ for $XYZ$ (unit cube), before both training and evaluation.
    In each experiment, methods are evaluated with mean $L_2$ Chamfer Distance (CD $\downarrow$) between the ground-truth set of surface points (normalised) and the surface points produced after rendering the particular method (Eq.~\ref{eq:CDdef}). 

        \begin{figure}[htbp]
            \centering
            \includegraphics[width=0.75\linewidth]{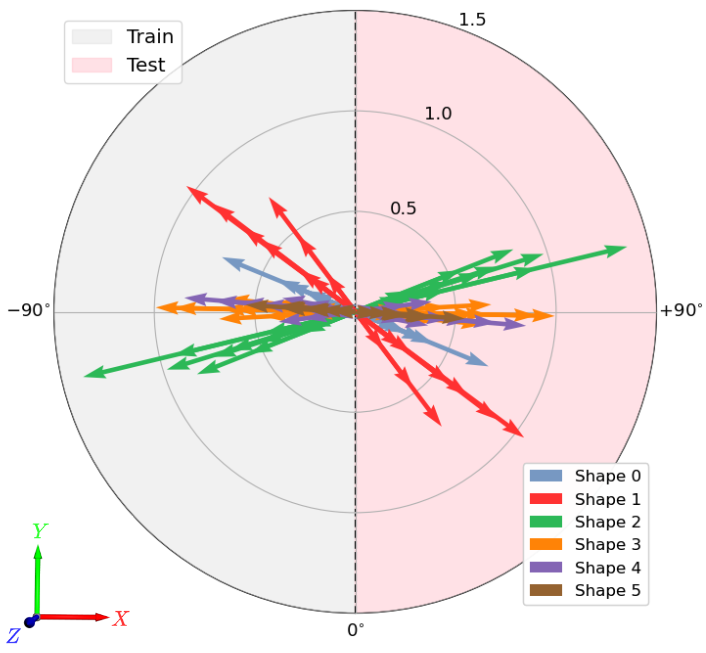}
            \caption{We evaluate generalisation performance on shapes deformed by forces from directions not seen in training. This figure shows the train and test split for each of the shapes in the \emph{Direction} experiment.}
            \label{fig:force_split_directions}
        \end{figure}

\begin{table}[b!]
	\centering
 	\caption{
        Chamfer Distance between shape reconstruction  and ground truth point cloud for nominal and deformed shapes seen in training.}

	\setlength{\tabcolsep}{5.0pt}
    \begin{tabular}{l cc}
        \toprule
        \multicolumn{1}{c}{ }                       &
        \multicolumn{2}{c}{Shape Reconstruction }    \\
        \multicolumn{1}{c}{ }                           &
        \thead{Nominal}             &
        \thead{Deformed}            \\
        \cmidrule(lr){2-3}
        VIRDO &
        $0.7505 \pm 0.5751$ &
        $0.7640 \pm 0.5890$ \\
        Ours &
        $\mathbf{0.2035} \pm \mathbf{0.05293}$ &
        $\mathbf{0.4940} \pm \mathbf{0.2868}$ \\
    	\bottomrule
	\end{tabular}

	\label{tab:shape-reconstruction}
\end{table}   

\subsection{Shape Reconstruction Results With Known Deformations}
    We report the normalised Chamfer Distance (CD $\downarrow$) on generated shapes from the withheld data as our metric for reconstruction quality between mesh reconstructions and ground truth point clouds, where CD results are multiplied by $\mathbf{10^{3}}$ for readability.
    Results show that our method outperforms a baseline method~\cite{Wi_2022} across all experiments by a large margin.
    Our method considers nominal shape samples to be a special case of a deformed shape, where a zero force vector has been applied at a random contact location, the \emph{trivial deformation}.
    For the initial experiment, training is conducted with the six nominal shapes (the \emph{trivial deformations} of each object), and the 24 deformed shapes for each object, and tested by reconstructing the shapes using query points seen in training.

    We validate our method by testing shape reconstructions on a known set of deformed objects and applied forces, results are shown in Table~\ref{tab:shape-reconstruction}. For \emph{nominal} shapes, \OurMethod{} recorded a mean normalised CD of 0.2035 for the test dataset, a 3.7$\times$ improvement over the baseline result of 0.7505 with a small variance for the CD values across the different shapes (standard deviation of 0.05293 compared to the baseline 0.5751). For \emph{deformed} shapes we recorded an average normalised CD of 0.4940 for the test dataset, a 1.5$\times$ improvement over the baseline result of 0.7640, again with lower standard deviation (0.2868 compared to 0.5890). These results show that our method can reproduce deformed shapes from training data with a higher level of detail than previous state-of-the-art methods.
    
  
    %
\subsection{Force Generalisation Results}
\label{subsec:force_gen}
    In Table~\ref{tab:force-generalisation} we designed four experiments to evaluate the force generalisation ability of \OurMethod{} against a baseline from \cite{Wi_2022}.
    In each experiment, a set of force-contact pairs is withheld from the training dataset.
    At test time, the model's performance is then evaluated by its ability to estimate deformed shapes resulting from the unseen force-contact pairs.
    In the first three experiments, a subset of five forces from the 24 deformation instances of each of the six objects were omitted, giving an 80/20 \% train-test split.
    
    In \emph{Random}, the five forces withheld are chosen at random to test the model's ability to generalise across the entire distribution of forces and to validate that the model hasn't memorised specific examples.
    For \emph{Lowest}, the five forces with the smallest magnitude $\|\ForceSym{}\|$ are withheld, to show the effect of subtle deformations from potentially underrepresented examples in training. In \emph{Highest}, the five greatest-magnitude forces are omitted from training to test the ability of the model to extrapolate beyond the training data.
    %

    For \emph{Direction}, the deformation samples are split 50/50 \% into the test/train datasets based on which side of the plane $y=0$ the constituent force vector points (Fig.~\ref{fig:force_split_directions}). 
    Discriminating direction instead of magnitude will appraise the model's cognisance of symmetry and whether it has sufficiently learnt to generalise to new directions/angles. 

\begin{figure}[htbp]
    \centering
    \includegraphics[width=1.0\linewidth]{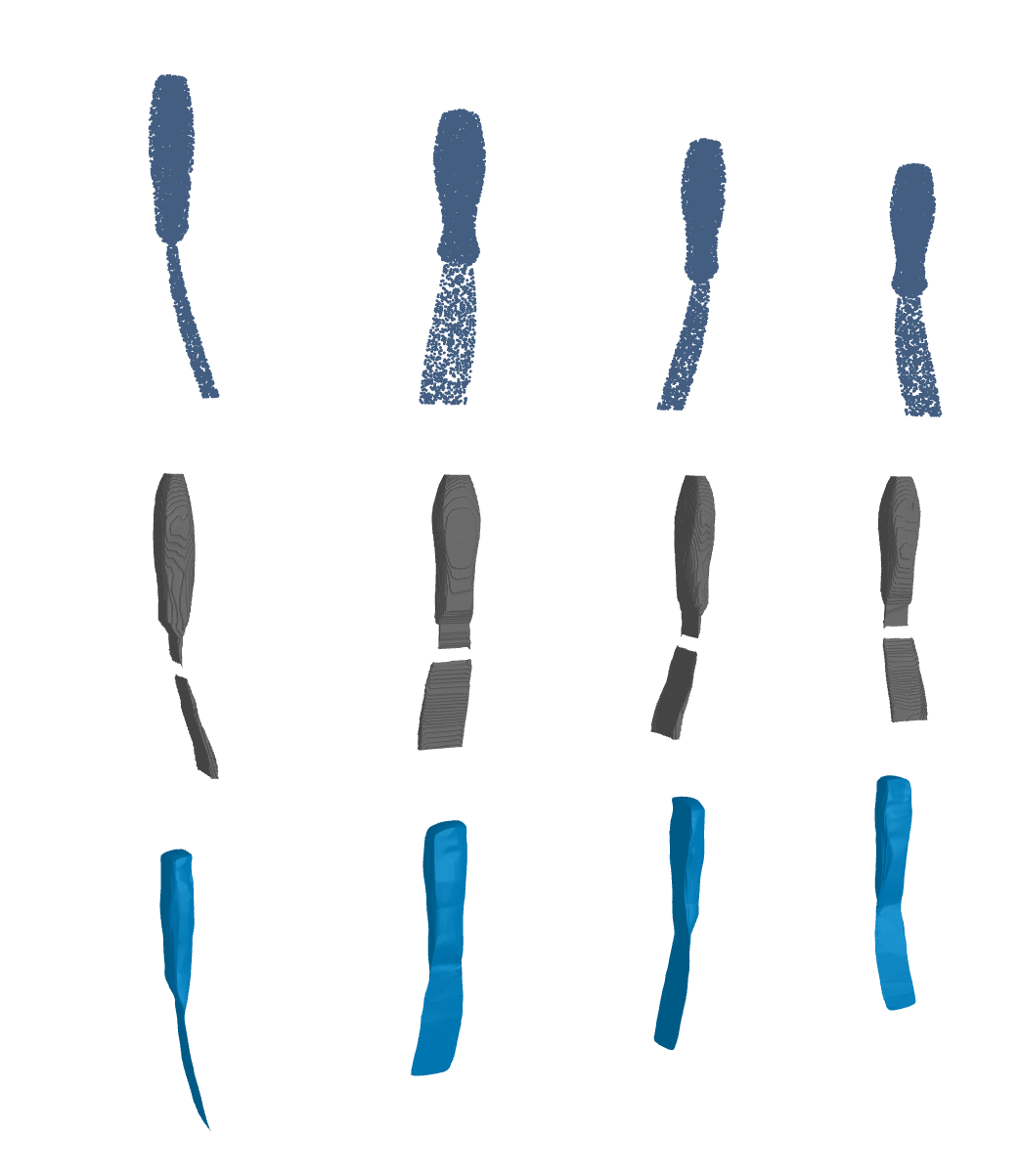}
    \caption{
    Evaluating the unseen: (shape 0, deform 0) in \ExperimentThree{}. 
    The top row is the ground truth point cloud.
    Mesh reconstructions from the next best method~\cite{Wi_2022} are shown in row 2, and Ours in row 3.
    Columns are of the same shape at different angles, for detailed inspection.
    }
    \label{fig:reconstructions}
\end{figure}

    Our method performs significantly better than the previous state-of-the-art~\cite{Wi_2022} when testing across the range of forces (\emph{Random}), with an average CD across all objects of 0.8720 compared to 300.5. Fig.~\ref{fig:reconstructions} shows reconstructions compared to ground truth point clouds (first row), and our reconstructions compared to~\cite{Wi_2022} (third and second row, respectively). In each of these examples, the applied forces were not provided in training, yet our method can generate accurate deformations.
    We highlight our method's attention to subtle changes in applied forces in experiments with low magnitude forces withheld (\emph{Lowest}) where our method can generate accurate reconstructions even from small forces (0.5793 compared to 0.9717). 

    Experiments withholding highest magnitude (\emph{Highest}) and forces from a single direction (\emph{Direction}) show that previous methods are particularly poor at generating deformations from forces not seen in the training data that require extrapolation, while our method produces meaningful reconstructions. \emph{Highest} shows a CD value of 1.993 compared to 375.8 for greater forces applied in testing. \emph{Direction} indicates that our method has learnt to understand the effect of forces applied from unseen orientations (5.895), whereas~\cite{Wi_2022} fails to produce meaningful reconstructions (362.1).
    \begin{table}[tp!]
	\centering
 	\caption{
        Chamfer Distance between shape reconstruction and ground truth point cloud for shapes deformed by forces that are unseen in training.
    }
	\setlength{\tabcolsep}{2.0pt} 
    \begin{tabular}{l cccc}
        \toprule
        \multicolumn{1}{c}{ }                       &
        \multicolumn{4}{c}{Force Generalisation}    \\
        \multicolumn{1}{c}{ }                           &
        \thead{Random}              &
        \thead{Lowest}              &
        \thead{Highest}             &
        \thead{Direction}           \\
        \cmidrule(lr){2-5}
        VIRDO &
        $300.5 \pm 111.80$  &
        $0.972 \pm 0.69$ &
        $375.8 \pm 82.10$  &
        $362.1 \pm 80.42$  \\
        Ours &
        $\mathbf{0.872} \pm \mathbf{0.34}$ &
        $\mathbf{0.579} \pm \mathbf{0.40}$ &
        $\mathbf{1.993} \pm \mathbf{1.21}$  &
        $\mathbf{5.895} \pm \mathbf{3.00}$  \\
    	\bottomrule
	\end{tabular}
	\label{tab:force-generalisation}
\end{table}
    %
    %
    %

\subsection{Object Generalisation Results}
\label{subsec:obj_gen}
    In the final set of experiments, we evaluate the model in a potential downstream use case by including a new shape with a reduced number of deformation examples. The experiments \emph{No deformed} and \emph{k deformed} further evaluate the force generalisation ability of our training method, from the perspective of data efficiency, by training the model with a reduced set of training samples.
    In the extreme case (\emph{No deformed}), we omit all 24 of the non-trivial deformations for a single object in training and only learn with the \emph{nominal} representation for this object, drawing on the deformation data for the other five objects to infer deformations. The test dataset comprises the 24 forces with corresponding deformations for the target object, giving the experiment an 84/16 train-test split.

    In the last experiment, (\emph{k deformed}), instead of removing all of the forces from the training set for the particular shape, we allow a varying number of $k$ randomly chosen non-trivial deformations to be learnt in training, to investigate the effect of increasingly adding training data as it becomes available, and to determine what amount of data would be required for a new object to yield a feasible deformation prediction in a downstream task.
    The training dataset will have $(5\times25) + 1 + k$ samples (5 existing shapes and all of their deformations, the nominal representation of the target shape, and k examples of its deformation), and the test dataset will have $25-k$ samples (unseen deformations from the target shape), 
    leading to a varying train/test split.
    %
    %
\begin{table}[htbp]
    \centering
    \caption{Each value of \textit{k} reflects how many deformed samples are in the training set, while average Chamfer Distance is reported for the remaining unseen deformations of a given target shape.}
    \setlength{\tabcolsep}{8pt} 
    \begin{tabular}{lcccc}
        \toprule
        \multicolumn{1}{c}{ } &
        \multicolumn{4}{c}{\textit{k Deformed}} \\
        \cmidrule(lr){2-5}
        \textit{k}   & 1 & 2 & 3 & 4  \\
        CD    & 4.75 & 0.521 & 0.429 & 0.383  \\
        \bottomrule
    \end{tabular}
    \label{tab:k-withheld}
\end{table}
%
%
\begin{figure}[htbp]
    \centering
    \includegraphics[width=0.28\linewidth, angle=90]{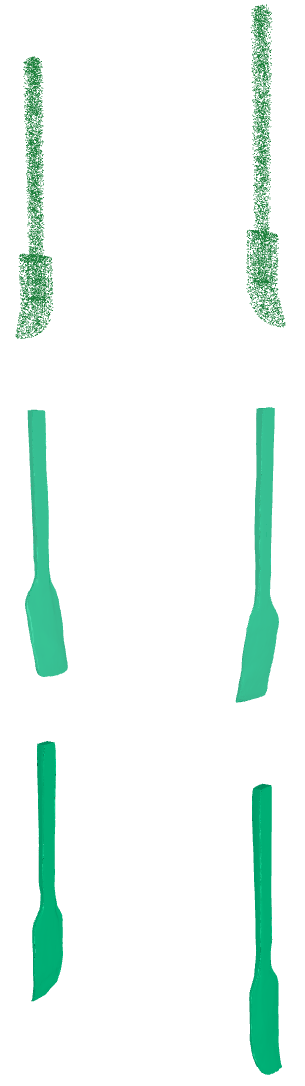}
    \caption{
    Evaluating the unseen: (shape 2, deform 48) not seen in the training data. The left-most column is the ground truth point cloud, middle and right columns are with $k=1$ and $k=2$ deforms introduced into the training set (\emph{k deformed}).
    }
    \label{fig:obj_gen1}
\end{figure}
%


    

%
%
\subsection{Model Efficiency}
%
For downstream robotics applications, real-world data collection is at a premium. These results show that we can adapt to a new object with unknown deformations and still produce useful reconstructions when a force is applied.
For the \emph{No deformed} experiments our method can reproduce deformed shapes with an average CD from 24 applied forces of 5.883, whereas comparative methods are unable to produce a watertight mesh for evaluation. Table.~\ref{tab:k-withheld} reports the average CD on reconstructions of all deformations of a target object in the \emph{k deformed} experiments. \textit{k} indicates the number of deformation examples provided in the training data.
These results demonstrate that our method improves reconstruction accuracy significantly over a range of unseen forces, generating accurate shapes (particularly in the deformed sections) in clear detail with a CD of 0.521 from only two training samples (Fig.~\ref{fig:obj_gen1}).

\begin{table}[htbp] 
    \centering
    \caption{Model parameters, training, and inference times.}
    \setlength{\tabcolsep}{8pt} 
    \begin{tabular}{lccc}
        \toprule
         \multicolumn{1}{c}{ } &
        \multicolumn{1}{c}{Parameters} &
        \multicolumn{1}{c}{Training (Hr)} &
        \multicolumn{1}{c}{Rendering (s)} \\
        \cmidrule(lr){2-4}
    
        VIRDO   & 52.6M & 8.779 & 5.0 \\
        Ours    & 1.0M  & 0.450 & 0.2 \\
        \bottomrule
    \end{tabular}
    \label{tab:model-efficiency}
\end{table}

Finally, we report results on model efficiency (Table.~\ref{tab:model-efficiency}), where our method is far more efficient in training (time and parameters), and can produce shape renderings that support real-time applications (0.2 seconds per reconstruction on a standard laptop with a NVIDIA A2000 GPU).
\section{Conclusion} \label{sec:conclusion}
We have presented a unified representation for encoding object deformations to create accurate and fine-grained reconstructions, making it suitable for downstream robotics applications.
Our method demonstrates strong generalisation to unseen forces and even broadening it's capability to unseen object, outperforming existing approaches in all results.
Furthermore, we show that the method is capable of real-time performance, and we plan to extend this work to robotic manipulation experiments in the future.


\balance{}

\bibliographystyle{./bibliography/IEEEtran}
\newpage
\bibliography{ref}



\end{document}